# DATA SCIENCE AS A NEW FRONTIER FOR DESIGN

Akın Osman Kazakçı
MINES ParisTech
akin.kazakci@mines-paristech.fr

The purpose of this paper is to contribute to the challenge of transferring know-how, theories and methods from design research to the design processes in information science and technologies. More specifically, we shall consider a domain, namely data-science, that is becoming rapidly a globally invested research and development axis with strong imperatives for innovation given the data deluge we are currently facing. We argue that, in order to rise to the data-related challenges that the society is facing, data-science initiatives should ensure a renewal of traditional research methodologies that are still largely based on trial-error processes depending on the talent and insights of a single (or a restricted group of) researchers.
It is our claim that design theories and methods can provide, at least to some extent, the much-needed framework. We will use a worldwide data-science challenge organized to study a technical problem in physics, namely the detection of Higgs boson, as a use case to demonstrate some of the ways in which design theory and methods can help in analyzing and shaping the innovation dynamics in such projects.

# 1 INTRODUCTION: DESIGN THEORY AND METHODS IN DATA-SCIENCE

10 years ago, the list of most valuable companies largely consisted of production or manufacturing companies that produced value from raw materials. Today, such lists contain more and more companies that produce value from information. What does this imply for design research and academics? Age old companies such as General Electric and Ford, despite all their might, are having difficulties in matching the impressive growth rate of companies such as Google, Amazon or Facebook. Undoubtedly, this growth has a direct relationship with innovation dynamics, rather than the sole introduction of some unprecedented technology (Tucker 2002; Gawer 2011). Assuming that systematic and repetitive innovation is driven by design (Le Masson *et al.* 2010; Verganti 2013; Brown 2014) a legitimate question to ask is: where does design theory and methods stand with respect to these transformations?

Indubitably, more than 50 years of research in design has allowed design research community to gather invaluable insights about the nature of creative activities. In a knowledge-based economy that is more and more driven by the creation of value from information, design theory and methods should be more present to sustain these creative activities. A look at the literature will tell, however, that this is not the case: a large majority of the current design research is still considering traditional products from engineering (mainly, electro-mechanical devices), architecture and product design. Many, if not all, theories and models suggested in design literature consider paradigms that are intimately tied to the core paradigm of the artifacts of a particular domain (such as functions, part-whole relationships). While it can be argued that these models are universal abstractions that can be applied on all variety of domains, no evidence exist, as far as the author is aware of, that they are being systematically used (or can be used to good effect) in the design processes in information science and technology.

The purpose of this paper is to contribute to the challenge of transferring know-how, theories and methods from design research to the design processes in information science and technologies. More specifically, we shall consider a domain, namely data-science, that is becoming rapidly a globally invested research and development axis with strong imperatives for innovation given the data deluge we are currently facing (Howe *et al.* 2008; Berriman and Groom 2011; Mattmann 2013). Data-science lies at the crossroads of computer science, machine learning and statistics. It can be argued that it is one of the core competencies of information science and technology companies such as LinkedIn, Twitter or Google. It is also equally important in scientific research given the rise of the fourth paradigm (Hey *et al.* 2010; Tolle *et al.* 2011). In this paper, we will use a data-science challenge organised to study a technical problem in physics as a use case to demonstrate some of the ways in which design theory and methods can help in analysing and shaping the innovation dynamics in such projects.

The plan of the paper is as follows. In Section 2, we present the background and current evolution of data-science landscape. We argue for a need of methodological renewal in data science practices. In Section 3, we present shortly the HiggsML challenge organised to gain insights into the study of Higgs boson in particle physics by means of machine learning algorithms. The challenge can be seen as crowdsourcing the development of algorithmic products, where participants try to improve upon existing results. In Section 4, we shall analyse some strategies used by the participants using a qualitative interpretative approach on the basis of available data combined with a design strategy modelling approach. In Section 5, we will propose and describe some alternative strategies generated using design models. Section 6 offers a brief conclusion.

# 2 DATA-SCIENCE AS A NEW PRIVILEGED SETTING FOR DESIGN: BACKGROUND AND CHALLENGES

Massive amounts of data are being produced in today's economic, scientific and social environment. Beyond infrastructure and engineering concerns, extraction of valuable information from data has become a paramount challenge for industry, academia and governments. The fact that the available data is growing exponentially requires pushing the limits of current analysis techniques towards radical breakthroughs. For instance, the Encyclopedia DNA Elements (ENCODE 2012) contains 15 TB of data (roughly the capacity of 30 thousand modern computers). By 2020, Square Kilometer Array (SKA) telescope will produce 1,5 billion times more data than that per year (Mattmann 2013).

Very recently, this imperative has given rise to a notion of *data science* in academic circles (Davenport and Patil 2012; Agarwal and Dhar 2014). Data science can be broadly defined as the design of automated methods to analyze massive and complex data in order to extract useful information. Data science and big data are closely related but not identical. Whereas big data covers a broad spectrum of themes on capturing, transferring, storing, searching, securely sharing, archiving, and analyzing massive data, the focus of data science is on the algorithmic and mathematical aspects of extracting new knowledge from data. As such, data science lies at the crossroads of computer science, applied mathematics, and statistics.

The academic debate on the notion of data science is being rapidly accompanied by institutional efforts and initiatives throughout the world. For example in US, following the announcement of the National Big Data R&D Initiative of the White House in 2012, both the national funding agencies (e.g., NSF, NIH, and DARPA) and universities engaged in large-scale top-down actions in order to promote data science and research on big data. The following cases are worth mentioning. The Research Data Alliance (RDA) is formed to accelerate data-driven innovation worldwide through research data sharing and exchange. New York University opened its Center for Data Science. University of Washington founded its eScience Institute. Berkeley launched its Institute for Data Science. The Moore and Sloan foundations announced a five-year 37.8M$ cross-institutional initiative to support the three previous institutes. In Europe, the University of Amsterdam announced the creation of its Data Science Research Center. Edinburgh University launched its Center for Doctoral training in Data Science. Imperial College London has associated with Zhejiang University to launch data science collaboration. In France, Université Paris-Saclay has created Centre for Data Science project oriented towards the analysis of scientific data.

The booming number of initiatives adds one more layer to the inherently difficult agenda of each data-science initiatives and programs: the necessity to distinguish themselves from others in the long run by means of scientific excellence and societal impact. Can data-science initiatives deliver the expected technical breakthroughs without methodological innovations? This paper argues that, in order to rise to the data-related challenges that are upon us, data-science initiatives should ensure a renewal of traditional research methodologies that are still largely based on trial-error processes depending on the talent and insights of a single (or a restricted group of) researchers.

It is our claim that design theories and methods can provide, at least to some extent, the much-needed framework. Approaches originating from design have been applied with great success in industrial settings for decades (Birkhofer 2011; Birkhofer *et al.* 2012; Agogué *et al.* 2014). Some of these theories and methods can be readily adapted or directly used in data-science. Design theory and methods can be used in two ways in data-science projects. First, they can be used to analyze and to make sense of the phenomenology underlying the process. In this context, these approaches will allow studying factors and mechanisms that are likely to affect the success. Second, they can help data-scientist in structuring the creative exploration processes in a controlled way. Given the overwhelming combination of design choices in even the simplest machine learning tasks, the use of a structuring method seems to be a must rather than option. In the following, we shall demonstrate by example how these two aspects can be combined.

## 3 HIGGSML DATA-CHALLENGE: CROWDSOURCING THE DESIGN OF ALGORITHMS

The Higgs boson machine learning challenge (HiggsML for short) has been organized by (Adam-Bourdarios *et al.* 2014) and hosted at Kaggle.com. The primary goal of HiggsML was to bring closer machine learning and physics communities, accelerating thus the transfer of advanced methods for data analysis from machine learning to physics. The presentation in this section is mainly based on (Adam-Bourdarios *et al.* 2014).

### 3.1 Scientific context and objectives

The Large Hadron Collider (LHC) began operating in 2009 after about 20 years of design and construction, and it will continue to operate for at least the next 10 years. The primary objective of LCL is to run experiments about particle physics. In 2012, two of its experiments, namely ATLAS and CMS, have claimed the discovery of the Higgs boson (Aad 2012; Chatrchyan 2012). François Englert and Peter Higgs, who theorized the existence and the role of this particle almost five decades ago, were given the Nobel Prize in Physics in 2013. This particle is seen as the final component of the Standard Model in particle physics describing relationships between subatomic particles and forces. The Higgs boson has been first observed in three distinct *decay channels* (that are all boson pairs). One of the next important scientific objectives is to study new decay channels (especially, the decay into fermion pairs, namely tau-leptons or *b*-quarks) and to precisely measure their characteristics. The first evidence of the *H* to tau tau channel was recently reported by the ATLAS experiment (Aad 2012). The HiggsML challenge has been put in place to use crowdsourcing in the development of learning algorithms that may improve on this analysis. Organizers expected that significant improvements were possible by re-visiting some of the *ad hoc* choices in the standard procedure, or by incorporating the objective function or a surrogate into the learning procedure design (Adam-Bourdarios *et al.* 2014).

### 3.2 Problem formulation for crowd-prototyping

The above scientific objective can be cast into the frame of a classification problem. Although some twists exist with respect to the traditional classification in machine learning, this main analysis process can be seen as separating as effectively as possible *events* as background or signal. Events are created by the Large Hadron Collider by provoking hundreds of millions of proton-proton collisions per second, creating large numbers of new particles. Online processing of these events discards most of the uninteresting ones. The remaining events (roughly four hundred per second) are then written on disks by a large CPU farm, producing petabytes of data per year for offline analysis. For the majority, these events correspond to known processes (called *background*): they are produced by the decay of particles that are rare in everyday terms, but known, since they have been discovered in previous experiments. The goal of the offline analysis is to find a (not necessarily connected) region in the feature space in which there is a significant excess of events (called *signal*) compared to what known background processes can explain (Adam-Bourdarios *et al.* 2014). Once the region has been fixed, a statistical (counting) test is applied to determine the significance of the excess. If the probability that the excess has been produced by background processes falls below a limit, the new particle is deemed to be discovered ($p = 2,87.10^7$ or its equivalent Gaussian significance $Z = 5$ sigma). The challenge has provided simulated data that can be used to prototype learning algorithms that can effectively separate signal from background. These algorithms are prototype in the sense that the objective is not to replace the techniques used by the physicists immediately, but rather to look for insights based on a crowdsourcing approach. In reality, any such insights and algorithms need to be adapted to the real problem that is more complicated than the version formulated for the HiggsML challenge (Adam-Bourdarios *et al.* 2014). Despite the similarity with ordinary classification problem, there are some important differences. The most important is that the formal objective function, the approximate median significance (AMS; see Figure 1) is quite different from the usual classification error objectives used in machine learning.

### 3.3 Results of the challenge: improvements in discovery significance

The HiggsML challenge has been a very successful event in many respects. A total of 1785 teams have participated to the challenge, which sets the record to date in all Kaggle competitions. The challenge provoked extensive exchanges between data-scientists and physicists, contributing to the primary goal of the challenge. Many participants with no background in physics have achieved considerable levels of success. More importantly, the discovery significance has been raised to ~3.8 sigma (up from 3.2 that can be achieved by a standard Boosted Decision Tree commonly used in physics). Although this is considered to be an important margin of improvement (roughly by 20%), this score is much lower than the target 5 sigma.

## 4 DESIGN DYNAMICS IN HIGGSML AND MACHINE LEARNING

### 4.1 Basic workflow in machine learning is a dominant design

While there is a considerable variety in applications of machine learning, there exist an extremely standardized general procedure that captures a majority of those. This basic workflow can be found in most, if not all, of the pedagogical material on introduction to machine learning. As such, it constitutes the pillar of the formation of data-scientists. It can be summarized as follows.

- **Select a classification method**
  A large variety of methods; some more general than others; most perform well on specific cases: the data scientist needs to know what method to use when.
- **Pre-process the data**
  Preparing the data for learning. Typical steps are normalization and replacing missing values.
- **Choose hyper-parameters**
  In most cases, some parameters need to be chosen by the data scientist. Methods such as grid search can be used to initialize these parameters.
- **Train**
  Fit the model to the data by minimizing some error metric; this is an optimization step (e.g. by stochastic gradient decent).
- **Monitor training performance**
  Traditionally, maximize prediction accuracy while avoiding over-fitting to the training data.
- **Return** $g(x): x \rightarrow i$**,** where $i$ is a class index

This workflow can be seen as a generic *process* design whose components should be decided at each level to reach a final design corresponding to a given application setting. In practice, no realistic application domain lends itself easily to this scheme without some adaptation.

This workflow can also be qualified as a dominant design (Abernathy and Utterback 1978). Rarely, in a majority of the (reported) machine learning and data-mining applications, it is changed in any way – more importantly, the data that is accessible, the algorithms that are available, and the problem formulations are all optimized to fit this generic design. From a pure design standpoint, this approach is similar to an old design strategy, dating back to (Redtenbacher 1852): a first (group of) designer(s) create a recipe, that the practitioner-designers adapt to their specific design context.

## 4.2 An analysis of design strategies used in HiggsML Challenge

### 4.2.1 Data and Methodology

In many data-science competitions, participants can communicate through a forum, present their approaches, comment on others' or ask for advice. Furthermore, it is a common practice to present commented code and tutorials, before or after the competition is over. These exchanges generate natural traces about what has happened. In HiggsML a total of 136 topics and 1401 posts are available at competition forums. Moreover, a subset of the posts contains links to participants' blogs or GitHub code repository, making available more material. In addition to those data, there exist challenge documentation and reference documents. Finally, at the end of the competition, a survey has been sent to participants by the organizers in order to collect more information about various aspects of their approaches. 35 teams, including many top participants, have answered the survey, providing valuable information (most of which are in unstructured text format).

Unfortunately, participants do not systematically report the approach used, nor the level of success. While available data provide a rich source of information on the participants activities and approaches, this information is not systematic, in the sense that, on any given topic, the point of view of a majority of participants are missing. Given the difficulty in quantifying such information and non-commensurability problems, our analysis is necessarily qualitative interpretative to major extent, combined with a design strategy modeling approach (Kazakci 2014) using C-K models (Hatchuel and Weil 2009). A limit of our approach is, thus, that whatever insight provided by our analysis needs to be confirmed by further study of the available data and by means of further experiments in a more controlled settings.

### 4.2.2 Dominant design as a trap leading to failure

Based on the available data, several C-K models representing competitors' design strategies have been built. Given the competition (and the forum) timeline, it was observed that most participants have been trapped in the traditional workflow, unable to figure out how to take into account AMS. Several participants reported that traditional performance metrics (such as accuracy, area under curve, …) did not yield good results for predicting signal and background points in a way to get AMS to acceptable levels.

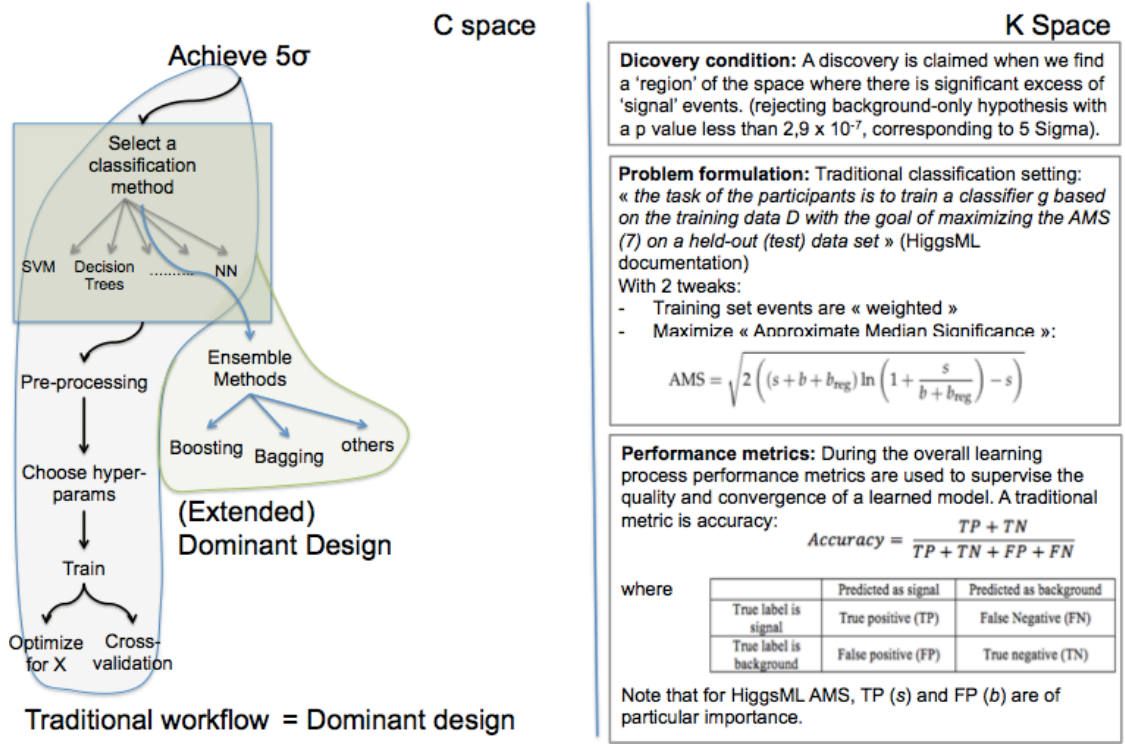

*Figure 1. Traditional machine learning strategy despite the innovative concept and associated objective function (AMS)*

### 4.3 Direct optimization of AMS

As previously mentioned, one of the expectations of the organizers was to see some participants develop strategies that successfully integrates AMS into the training procedure. Available data captures only a handful of attempts, as mapped in Figure 2. One such consideration is the integration of AMS into the gradient boosting algorithm. This algorithm is a variant of the gradient descent optimization technique, which can accommodate any function as an objective function, as long as it is derivable. Another consideration is to use AMS during with random (decision) forest type approaches. These approaches build decision trees on a randomly selected subset of the data where trees are grown by splitting nodes in a way that optimize the fit of the tree with the subset of the data. Both of the attempts mentioned in the forums did not succeed. Note that, this does not invalidate the overall concept. A third approach was proposed by (Mackey and Bryan 2014). They have realized that the AMS function is a particular case of more general form for a class of utility functions. By applying knowledge from convex optimization domain (namely, Fenchel-Young inequality), they were able to obtain theoretical results on AMS. Based on their result, they were able to design an original procedure, called weighted classification cascade that alternates between an optimization step and a learning step. From a design standpoint, this is a genuine conceptual expansion; not only it partitions the dominant design in an unexpected way, it does so by activating a new and untapped knowledge pocket. Unfortunately, the corresponding approach has not performed well (ranked 461th). However, once again, it cannot be claimed that the overall design path can be invalidated – a more systematic study of this expansion seems worthwhile.

### 4.4 The path of the winners: statistical efficiency rather than AMS

Most successful entries (AMS >3.7) have been obtained by exploiting strategies that are different than both the traditional workflow and the direct optimization of AMS paths. In fact, top competitors have treated the optimization of AMS as an external issue and focused on the quality of the learning process. In most cases, this consists in replacing traditional performance measures by some cross-validation strategy and used an ensemble of learners (an approach known for reducing predictive variance, thus achieving more robustness in predictions) – both of which are common practices in machine learning.

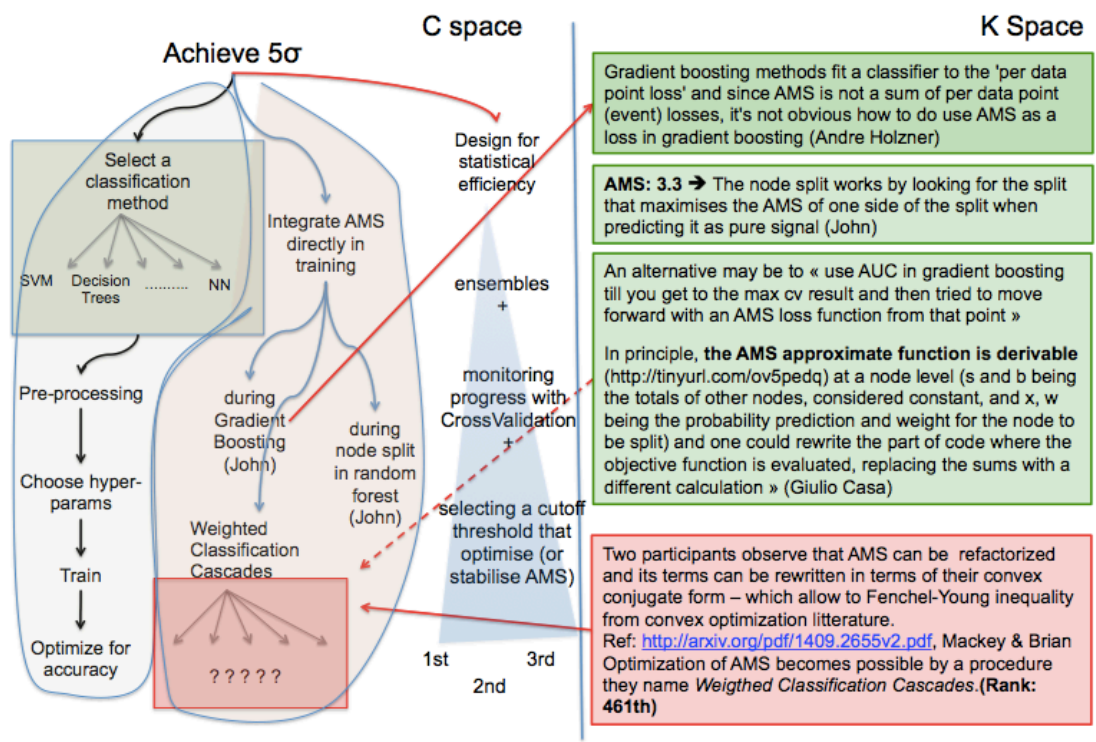

*Figure 2. Direct integration of AMS into the learning process – some short but fruitful explorations*

### 4.5 Public guides facilitating public fixations: Two particular cases

During the competition, the interactions on the forums have shaped the way participants develop and deploy their strategies. Based on the votes for and the number of replies to various topics in competition forums, it can be observed that two events have been particularly salient. First, a group of physicists has released two new variables mid-competition (193 votes, 107 replies and 7147 views on forums). These variables can be used as inputs to the learning algorithms (they also provided code to compute these values). The physicists expected that experts in machine learning could make a better use of these handcrafted features. Looking at the post-competition survey, it cannot be stated they were necessary, since there exist many successful entries that did not make use of them.

A second, and probably more important event corresponds to a public guide for getting the AMS up to 3.6. Mid-competition, one of the participants provided a guide and the associated code to promote his software package. Several participants simply copied the code to level their scores to 3.6. So far, the topic received 102 votes, 97 replies and 29976 views. Given the number of total participants is 1785, these numbers indicate a high level of attention paid to the proposed suggestion. Although several participants have replicated the solution, very few participants got above 3.7 with it.

### 5  BROADENING THE SCOPE OF EXPLORATION BY DESIGN

This section reports on an ongoing investigation we conduct with some of the HiggsML organization committee members about the use of design theories and models in the study of Higgs boson by means of machine learning techniques. Our focus is on generation and mapping of alternative approaches to the problem, with an emphasis on techniques that have not been used so far. Our objective is to broaden the scope of problem formulation. To this end, we have looked into some of the problem characteristics and conducted a literature survey based on those characteristics. In fact, the HiggsML challenge has several characteristics (severely imbalanced signal to background ratio, class skewness, small disjuncts, rare classes, class overlap, noise, systematic errors…) that point to various knowledge domains in the machine learning literature. To build an initial C-K model, we choose the class imbalance problem (CIP); a partial view of the concept space of our model is depicted in Figure 3. The base concept can be stated as "*Achieve 5 sigma by dealing with class imbalance problem*".

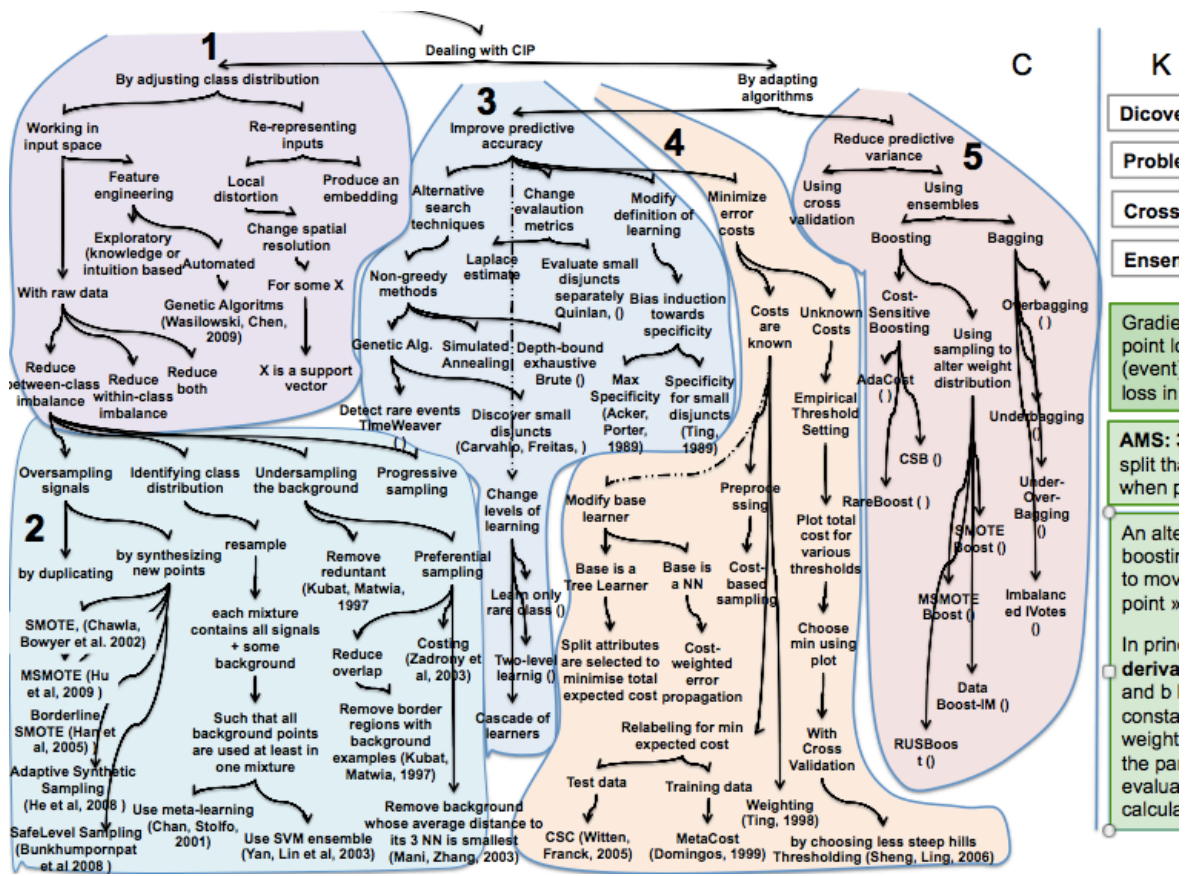

*Figure 3. A concept space representation for dealing with class imbalance problem*

As far as we can tell from the available data, despite the class imbalance being one of the salient characteristics of the problem and the existence of specialized literature on the topic, the participants have not considered CIP explicitly. We found one mention of CIP on the forums, whose authors stated that the AUC metric was not a reliable metric for this competition due to the class imbalance problem. But, they did not follow up to rectify this shortcoming or to propose alternative approaches, contrary to the specialized literature on the topic.

The final model is too detailed to be fully presented here. Instead we shall focus on the general structure of the model, some suggestions we can make based on the model and the overall process of modeling such an innovation field. The general structure of the model is derived from two basic partitions; we can deal with CIP either by $C_1$: *adjusting the class distributions* or by $C_2$: *adapting the learning algorithms*. For $C_1$, the most explored design path is sampling strategies (marked with 2 in Figure 3), which we shall come back to later on. Under $C_1$ (marked with 1 in Figure 3), we can also find other general strategies of pre-processing data such as feature engineering. For example, (Wasikowski and Chen 2010) used genetic algorithms to engineer some features in the case of class imbalance and tested the effect of several metrics for obtaining good feature sets. There are several strands of alternative approaches pertaining to $C_2$ (marked with 3, 4 and 5 in Figure 3). A large part of those can be categorized under $C_3$: *Improve predictive accuracy*. We distinguish $C_{34}$: *Minimizing error costs* (marked with 4 in Figure 3), since some concepts under this branch are conceptually related to parts of the winning approaches in HiggsML challenge. Under $C_4$: *Reducing predictive variance*, we have regrouped mainly approaches that use ensemble of learners adapted to CIP in some way (mostly, combining other strategies from 2 and 4). In the model, a wealth of ideas some of which are indeed surprising exist (in the sense that they are far from traditional concepts). For the purposes of this paper, we are going to briefly develop some of them.

- A quite original idea under $C_1$ is the class boundary alignment (Wu and Chang 2003). The idea consists in applying a local distortion to the data representation only around boundaries separating signal and background classes. In their cases, this is done by "conformally spreading" the area around the class-boundary outward by applying and adjusting a Riemannian metric on the kernel

- function of a support vector machine (SVM). Thus, their approach is specific to SVMs and needs to be adapted to other (and more powerful) learning algorithms in the HiggsML case.
- Under $C_1$, the bulk of the work has been on sampling strategies to re-balance the class distribution. While there exist works claiming that within-class imbalance has to be maintained (Nickerson *et al.* 2001), most work focus on between-class imbalance. The line of works that stands out the most consists in *oversampling* the rare class (signal) by creating synthetic data points. In the case of SMOTE (Chawla *et al.* 2002), new minority class examples are created by interpolating a few minority class instances that are close to each other for oversampling the training set. As far as we know, no HiggsML participants have used such a technique or whether it would be useful or acceptable to use such a strategy with real CERN data.
- Another conceptual path that is not reported by the participants is the idea of progressively sampling for active learning (Ertekin *et al.* 2007a; Ertekin *et al.* 2007b). The primary aim of active learning is to ask for class labels (instead of receiving some particular subset beforehand) only for the data points that would accelerate learning the most. This is an intriguing idea in the study of Higgs (but also for the challenge) since this approach targets explicitly unknown and undecidable areas (e.g. where background and signal cannot be properly separated).
- Under $C_3$, a number of conceptual paths exist. First, there are suggestions to use alternative search techniques for model fitting. Traditional approaches use some form of greedy search. Variations of other optimization techniques such as genetic algorithms or simulated annealing can be used to alleviate problems due to class imbalance in this setting. Second, alternative performance metrics specifically designed for CIP can be used. Literature suggests, for instance, the use of Laplace estimates or the evaluation of small disjuncts separately. Third, the definition of learning itself, mainly consisting in privileging generalization can be altered – to favor specificity of the rare class (in HiggsML, the signal class).
- Under $C_{34}$, we find approaches using the notion of cost for different kinds of error. In HiggsML setting this may be interesting since the true positives (correctly predicted signal) and false positives (incorrect background particles predicted as signal) do not have same costs. A particular idea we have discovered through modeling (and not through our literature review) is that the asymmetric costs can be reflected to errors back-propagation in neural networks (i.e. penalizing more particular types of errors). Finally, when the costs are not known, approaches have been developed to experimentally determine a ratio between error components by setting a threshold – this corresponds to the central approach used by top competitors.

## 5.1 A note on C-K modeling

It should be noted that most of the concepts presented in our model awaits validation since they have been suggested for small databases or much less important imbalance ratio settings. Furthermore, let us note that the concept tree is not exhaustive: not only this is unlikely to be possible, it is not needed either. Our aim being the demonstration of how design models can help the generation of alternatives in data-science, the current model illustrates our propos. It should also be remarked that this tree has not been generated (almost) automatically from a literature review. There exist several discrepancies in the literature connecting various approaches and the modeler needs to fill in those gaps in a way to build a comprehension of the domain for oneself. Also, the tree is not simply a taxonomy since it does not correspond to a unique hierarchy; some concepts may as well have appeared elsewhere in the tree. This model is not meant to be « final » in any sense, not anymore than any reasoning process would be: new opportunities or challenges might question the presented model and require that it evolves. This evolution is likely as the model will be considered shortly with the organizers and the winner of the challenge in an upcoming series of creative design workshops and programming bootcamps.

## 6 CONCLUSION: DESIGN IS EVERYWHERE... SO SHOULD BE DESIGN RESEARCH

At the hart of design theory and methods lies the notion of controlled creativity. To deal with the inherent complexity of its creative design problems, data science needs renew its methodologies. We have presented a case illustrating in which way design theory and methods can help. On the other hand, it is readily acknowledged in design research community that design is everywhere. This paper is an invitation to get design research in non-traditional areas of challenge the society faces today.

**Acknowledgements**. The author is grateful to Balázs Kégl for countless hours of discussions about the HiggsML and machine learning in general, and to Alexandre Gramfort for his comments.